\documentclass[runningheads,a4paper]{llncs}
\usepackage{amssymb}
\usepackage{graphicx}
\usepackage{cite}
\usepackage{amsmath,amssymb,amsfonts}
\usepackage{algorithmic}
\usepackage{adjustbox}
\usepackage{graphicx}
\usepackage{multirow}
\usepackage{textcomp}
\usepackage{lastpage}
\usepackage{ctable}
\usepackage{fancyhdr}
\usepackage{color}
\usepackage[multiple]{footmisc}

\usepackage[font=small,labelfont=bf]{caption}

\newcommand{\rom}[1]{\uppercase\expandafter{\romannumeral #1\relax}}

\title{Dynamic voting in multi-view learning for radiomics applications}
\author{Hongliu CAO\inst{1,2} \and
	Simon Bernard\inst{2} \and Laurent Heutte\inst{2}\and Robert Sabourin\inst{1}}
\author{Hongliu Cao\inst{1,2}%
	\thanks{Corresponding author, email: caohongliu@gmail.com}%
	\and
	Simon Bernard\inst{2} \and Laurent Heutte\inst{2}\and Robert Sabourin\inst{1}}
\authorrunning{Hongliu CAO et al.} 
\institute{LIVIA, \'Ecole de Technologie Sup\'erieure,  Universit\'e du Qu\'ebec, Montreal, Canada,\\
	\and
	Normandie Univ, UNIROUEN, UNIHAVRE, INSA Rouen, LITIS, Rouen, France}
\begin{document}
	\maketitle
	\begin{abstract}
		Cancer diagnosis and treatment often require a personalized analysis for each patient nowadays, due to the heterogeneity among the different types of tumor and among patients. Radiomics is a recent medical imaging field that has shown during the past few years to be promising for achieving this personalization. However,  a recent study shows that most of the state-of-the-art works in Radiomics fail to identify this problem as a multi-view learning task and that multi-view learning techniques are generally more efficient. In this work, we propose to further investigate the potential of one family of multi-view learning methods based on Multiple Classifier Systems where one classifier is learnt on each view and all classifiers are combined afterwards. In particular, we propose a random forest based dynamic weighted voting scheme, which personalizes the combination of views for each new patient to classify. The proposed method is validated on several real-world Radiomics problems.

	\end{abstract}
	
	\section{Introduction}
	
	One of the biggest challenges of cancer treatment is the inter-tumor heterogeneity and intra-tumor heterogeneity. It demands for more personalized treatment. In Radiomics, a large amount of features from standard-of-care images obtained with CT (computed tomography), PET (positron emission tomography) or MRI (magnetic resonance imaging) are extracted to help the diagnosis, prediction or prognosis of cancer \cite{cao2018dissimilarity}. Many medical image studies like \cite{sorensen2010quantitative,sluimer2006computer} have already tried to use quantitative analysis before the existence of Radiomics. However, with the development of medical imaging technology and more and more available softwares allowing for more quantification
	and standardization, Radiomics focuses on improvements of image analysis, using an automated high-throughput extraction of
	large amounts of quantitative features \cite{lambin2012radiomics}.
	Radiomics has the advantage of using more useful information to make optimal treatment decisions (personalized medicine) and make cancer treatment more effective and less expensive \cite{kumar2012radiomics}. 
	
	Radiomics is a promising research field for oncology, but it is also a challenging machine learning task. In the work \cite{cao2018dissimilarity}, the authors identify Radiomics as a challenge in machine learning for the three following reasons:
	\textbf{(i) small sample size}: due to the difficulty in data sharing, most of Radiomics data sets have no more than 200 patients; 
	\textbf{(ii) high dimensional feature space}: the feature space for Radiomics data is always very high dimensional compared to the sample size; 
	\textbf{(iii) multiple feature groups}: different sources and different feature extractors are used in Radiomics - the most used features include tumor intensity, shape, texture, and so on\cite{aerts2014decoding} - and it may be hard to exploit the complementary information brought by these different views \cite{cao2018dissimilarity}.

	When the three challenges are encountered in a classification task, it can be seen as an HDLSS (High dimension low sample size) Multi-View learning task. Now most studies in Radiomics ignore the third challenge and propose to simply concatenate different feature groups and to use a feature selection method to reduce the dimension. However, a lot of useful information may be lost when only a small subset of features is retained \cite{cao2018dissimilarity}, and the complementary information that different feature groups can offer may be ignored \cite{cao2018improve}.
	
	In contrast to the current studies that treat Radiomics data as a single-view machine learning task, we have proposed in our previsous work to cope with Radiomics complexity using an HDLSS multi-view paradigm \cite{cao2018dissimilarity}: we have used a naive MCS (Multiple Classifier Systems) based method which turns out to work well for Radiomics data but not significantly better than the state of the art methods used in Radiomics. Here we want to further investigate the potential of the MCS multi-view approach. Hence we propose several less simplistic MCS based methods including static voting and dynamic voting methods to combine classification results from different views. Our main contribution in this paper is thus to propose a new dynamic voting scheme to give a personalized diagnosis (decision) from Radiomics data. This dynamic voting method is designed for small sample sized dataset like Radiomics data and uses a large number of trees in random forest to provide OOB (Out Of Bag) samples to replace the validation dataset. 
	
	The remainder of this paper is organized as follows. Related works in Radiomics and multi-view learning are discussed in Section 2. In section 3, the proposed dynamic voting solution is introduced.  Before turning to the result analysis (Section 5), we describe the data sets chosen in this study and provide the protocol of our experimental method in Section 4. We conclude and give some future works in Section 6.
	
	\section{Related Works}
	In the state of the art of Radiomics, groups of features are most often concatenated into a single feature vector, which results in an HDLSS machine learning problem. In order to reduce the high dimensionality, some feature selection methods are used : in the work of \cite{parmar2015radiomic} and \cite{aerts2014decoding}, they used feature stability as a criterion for feature selection While in the work of \cite{song2016non}, they used a SVM (Support Vector Machine) classifier as a criterion to evaluate the predictive value of each feature for pathology and TNM clinical stage. Different filter feature selection methods have also been compared along with reliable machine learning methods to find the optimal combination \cite{parmar2015radiomic}. Generally speaking, the embedded feature selection method SVMRFE shows good performance on different Radiomics applications \cite{cao2018dissimilarity}. 
	
	A lot of studies have been done on multi-view learning and according to the work of \cite{serra2015mvda}, there are three main kinds of solutions: early integration, intermediate integration and late integration. 
	Early integration concatenates information from different views together and treats it as a single-view learning task\cite{serra2015mvda}. The Radiomics solutions discussed above all belong to this category.
	Intermediate integration combines the information from different views at the feature level to form a joint feature space. Late integration method firstly builds individual models based on separate views and then combines these models. Compared to intermediate and late integration methods, early integration always leads to high dimensional problems and the feature selection methods used in the state of the art of Radiomics can easily filter a lot of useful information.

	In \cite{cao2018dissimilarity}, MCS based late integration methods (with simple majority voting) have shown a big potential and a lot of flexibility on Radiomics data.  In this work, to further investigate the potential of MCS for Radiomics applications, both static and dynamic combinations are tested. The intuition behind static weighted voting is that different views have different importances for a classification task. While the intuition behind proposing dynamic voting methods is that, due to the heterogeneity among patients, different patients may rely on different information sources. For example, for a patient A, there may be more useful information in one view (e.g. texture or shape features) while for a patient B, there may be more useful information in another view (e.g. intensity or wavelet features).
	Three dynamic integration methods were considered in the work of \cite{tsymbal2006dynamic}: DS (Dynamic Selection), DV (Dynamic Voting), and  DVS (Dynamic Voting with Selection). The difficulty in multi view combination is that the number of views is fixed and usually very small. In this case, dynamic selection methods may not be applicable. Hence, we focus on dynamic voting method in this work. However, traditional dynamic voting  methods demand a validation dataset \cite{cruz2018dynamic}. In Radiomics, the data size is too small to have a validation dataset. In the next section, we propose a dynamic voting method based on the random forest dissimilarity measure and the Out-Of-Bag (OOB) measure, without the need of validation dataset.
	
	\section{Proposed MCS based solutions}
	As explained in the Introduction, the simple MCS based late integration method used in \cite{cao2018dissimilarity} has shown a good potential for Radiomics. In this section, we use several more intelligent voting methods including static voting and dynamic voting to test if they can get significantly better.

	For multi-view learning tasks, the training set $\mathbf{T}$ is composed of $Q$ views: $\mathbf{T}^{(q)} = \{(\mathbf{X}_1^{(q)}, y_1),\dots,(\mathbf{X}_N^{(q)}, y_N)\}, q=1..Q$. 
	Generally speaking, the MCS based late integration method builds a classifier $C^{(q)}$ for each view $\mathbf{T}^{(q)}$. During test time, for each test data $\mathbf{X}_t$, $C^{(q)}$ will predict the class label $label_t^{(q)}$ of $\mathbf{X}_t$. Finally, the predicted labels from all the views $\{ label_t^{(1)}, label_t^{(2)}, \dots,label_t^{(Q)}\}$ can be combined either by majority voting or weighted voting.
	
	Here Random forest is chosen as the classifier for each view $\mathbf{T}^{(q)}$ because it can deal well with different data types,  mixed variables and high dimensional data \cite{cao2018dissimilarity}. Random forest can also offer the OOB measure, which can be used as a measure for static weight and also to replace extra validation dataset for dynamic voting methods. In addition, random forest also provides a proximity measure, which can be used to calculate the neighborhood of a test sample\cite{tsymbal2008dynamic}. 
	
	Firstly, for each view $q$, a Random Forest $\mathbf{H}^{(q)}$ is built with $M$ decision trees, and is denoted as in Equation \eqref{e2}:
	\begin{equation}\label{e2}
	\mathbf{H}(\mathbf{X}) = \{h_k(\mathbf{X}),k=1,\dots,M\}
	\end{equation}
	where $h_k(\mathbf{X})$ is a random tree grown using bagging and random feature selection. We refer the reader to \cite{breiman2001random,biau2016random} for more details about this procedure.

	For a $J$-class problem with $label_t^{(q)}=i$, where $i\in \{1,2,\dots,J \}$, a weight $W^{(q)}$ is used for each view $q$ (for the case of majority voting, all $W^{(q)}=1$). The final decision is made by:
	\begin{equation}\label{cap1}
	y_{t} = \underset{j\in \{1,2,\dots,J \}}{Max}(\sum_{q=1}^{Q}I(label_t^{(q)}=j)\times W^{(q)})
	\end{equation}
	$I()$ is an indicator function, which equals to 1 when the condition in the parenthesis is fulfilled and 0 otherwise. 

	\subsection{ WRF (Static Weighted Voting)}

	To calculate the weights for static voting, we need a measure to reflect the importance of each view to give a final decision. Usually, the prediction accuracy over a validation dataset can be used for that. However, Radiomics data have very small sample size, and it is impossible to have extra validation data. Hence we propose to use the OOB accuracy of each random forest $\mathbf{H}^{(q)}$ as the static weight $W^{(q)} $ for each view:
	\begin{equation}\label{capstatic}
	W_{static}^{(q)} =OOB_{accuracy}(\mathbf{H}^{(q)})
	\end{equation}
	
	When Bagging is used in a random forest, each bootstrap sample used to learn a single tree is typically a subset of the initial training set. This means that some of the training instances are not used in each bootstrap sample (37\% in average; see \cite{breiman1996out} for more details). For a given decision tree of the forest, these instances, called the Out-of-bag (OOB) samples, can be used to estimate its accuracy.
	To use OOB to measure the accuracy of a random forest, the concept of sub-forest is used. When the forest size is big, all training data have a high probability to be an OOB sample at least once. Hence, for each OOB sample $\mathbf{X}_{OOB}$, the trees that did not use this data as training sample are grouped together as a sub-forest $\mathbf{H}_{sub(\mathbf{X}_{OOB})}$ (which can be seen as a representative of the complete random forest $\mathbf{H}$) to give a prediction on $\mathbf{X}_{OOB}$. The overall accuracy of the sub-forests predictions on all OOB samples is then used as OOB accuracy for a random forest $\mathbf{H}$.  We refer the reader to the work of \cite{breiman1996out} for further information about OOB measure.
	
	\subsection{GDV (Global Dynamic Voting)}
	In static voting, we believe that different views have different importances for classification. However, with dynamic voting, we can personalize this importance with an assumption that the importances of views are different for different patients. One easy access to this kind of "personalized" information is the prediction probability of each test sample as it shows generally how confident the classifier $C^{q}$ is on the test data. 
	
	The predicted class probabilities of a test sample $\textbf{X}_t$ for random forest are computed as the mean predicted class probabilities of the trees in the forest. The class probabilities of a single tree is the fraction of samples of the same class in a leaf.
	The global weight $W_{global}^{(q)}$ of view $q$ for each test data $\textbf{X}_t$ is simply the predicted probability (posterior probability obtained from $\mathbf{H}^{(q)}$) for the most confident class of random forest, which measures the overall confidence rate of label prediction based on all the training data:
	\begin{equation}\label{lac}
	W_{global}^{(q)} = P(label_t^{(q)} \mid \mathbf{X}_t,\mathbf{H}^{(q)})
	\end{equation}
	
	$W_{global}^{(q)}$ generally reflects how confident the classifier $\mathbf{H}^{(q)}$ is when predicting the label of a test sample.  But it also means the global measure is not very personalized. To capture more personalized information, we propose in the next subsection the local weight measure. 
	\subsection{LDV (Local Dynamic Voting)}
	A local weight usually means the performance or confidence of a classifier in a smaller neighborhood in validation data of a test sample. It usually demands two measures: firstly, a distance measure to find the neighborhood; secondly the competence measure to evaluate the performance of the classifier in the neighborhood.
	RFD (random forest dissimilarity) in this work is used as a distance measure to find the neighborhood of a given test sample, while OOB measure is used to replace the validation dataset. 
	
	The RFD measure $\mathbf{D}_{\mathbf{H}}$ is inferred from a RF classifier $\mathbf{H}$, learned from training data $\mathbf{T}$. For each tree in the forest, if two samples end in the same terminal node, their dissimilarity is 0 otherwise 1. This process goes over all trees in the forest, and the average value is the RFD value (more details are given in \cite{cao2018dissimilarity}). 
	It can be told that compared to other dissimilarity measures, RFD takes the advantage of class information to measure the distance \cite{cao2018dissimilarity}. 
	
	To calculate the local weight $W_{local}^{(q)}$, RFD is used to find the neighborhood $\theta_{\textbf{X}}$ of each test instance $\textbf{X}$ by choosing the most $n_{neighbor}$ similar instances in training data.   
	The OOB measure over $\theta_{\textbf{X}}$ is then used to calculate the local weight.  Unlike in the work of \cite{tsymbal2006dynamic} using OOB to measure the individual tree accuracy, here OOB is used to measure the performance of the RF classifier. With $\theta_{\textbf{X}}$, the local weight can be easily calculated with OOB measure:
	\begin{equation}\label{lc}
	W_{local}^{(q)} =OOB_{accuracy}(\mathbf{H}^{(q)},\theta_{\textbf{X}})
	\end{equation}
	
	The idea of local weight here is similar to OLA (Overall Local Accuracy) used in dynamic selection \cite{cruz2018dynamic}. There are two main differences: firstly, LDV uses the random forest dissimilarity as a distance measure which carries both feature information and class label information while OLA uses Euclidean distance which may suffer from the concentration of pairwise distance \cite{aggarwal2001surprising} in high dimensional space; secondly, OLA requires a validation dataset while LDV does not. 
	
	\subsection{GLDV (Global\&Local Dynamic Voting)}
	From the previous two subsections, we can see that $W_{global}^{(q)}$ uses global information from all training data and measures the confidence of the classifier. But it has also the risk of being too generalized and lacks of personalized information. On the other hand, $W_{local}^{(q)}$ uses information on the neighborhood of the test sample to give a more personalized measure which can better represent the heterogeneity among cancer patients  but may lose the global vision at the same time. Hence we propose a measure that takes both measures into account. 
	
	With each $\mathbf{H}^{(q)}$, the global weight $W_{global}^{(q)}$ and the local weight $W_{local}^{(q)}$ are calculated respectively and the combined weight ${W}^{(q)}_{GL}$ is calculated by taking advantage of both global and local information together:
	\begin{equation}\label{dw}
	{W}^{(q)}_{GL} =W_{global}^{(q)}\times W_{local}^{(q)}
	\end{equation}

	The reason why we choose to multiply global weight and local weight for deriving a combined weight, is that, as it is explained previously, $W_{global}$ lacks personalized information, but it can be counter-balanced by $W_{local}$ to give more preference in some situations. For example, when ${W}^{(q)}_{global}$ agrees with $ {W}^{(q)}_{local}$ on a particular view $q$, if both weights are small, then ${W}^{(q)}_{GL}$ becomes even smaller as we do not have confidence on this view; if both weights get bigger and bigger, then ${W}^{(q)}_{GL}$ gets closer and closer to both weights, especially local weight. On the contrary, when $ {W}^{(q)}_{global}$ disagrees with $ {W}^{(q)}_{local}$, it is hard to make a decision with a disagreement (as we need prior knowledge to decide to choose global or local weight); hence we penalize ${W}^{(q)}_{GL}$ as long as there is a disagreement (${W}^{(q)}_{GL}$ is smaller than 0.5) but still with a preference to $ {W}^{(q)}_{local}$.

	\section{EXPERIMENTS}
	
	In this study, we use several publicly available Radiomics datasets. A general description of all datasets can be found in Table \ref{tab:data} where $IR$ stands for the imbalance ratio of the dataset. More details about these datasets can be found in the work of \cite{zhou2017mri}.

			\begin{center}
				\begin{adjustbox}{max width=1\textwidth}
					
					\begin{tabular}{c|ccccc}
						\hline
						& \#features & \#samples & \#views & \#classes & IR \\
						\hline
						nonIDH1  & 6746 &84 & 5 & 2&3 \\
						\hline
						IDHcodel & 6746 &67 & 5 & 2&2.94 \\
						\hline
						lowGrade & 6746 &75 & 5 & 2&1.4 \\
						\hline
						progression & 6746 &75 & 5 & 2&1.68 \\
						\hline
					\end{tabular}
				\end{adjustbox}
				\captionof{table}{Overview of each dataset.}
				\label{tab:data}   
			\end{center}

	The main objective of the experiment is to compare the state of the art Radiomics methods to static and dynamic voting methods. In total six methods are compared: one state of the art Radiomics method, i.e. SVMRFE; two static weighting methods, i.e. MVRF (combines RF results with majority voting as in \cite{cao2018dissimilarity}) and WRF (combines RF results with weights as in Section 3.1, the weights are the OOB accuracy of each $\textbf{H}^{(q)}$);
	three dynamic weighted voting methods, i.e. GDV, LDV and GLDV as described in the previous section. 

	For the two dynamic voting methods that use local weights, LDV and GLDV, the neighborhood size $n_{neighbor}$ is set to 7 according to the work of \cite{cruz2018dynamic}. For SVMRFE, the number of selected features is defined as in \cite{cao2018dissimilarity} according to the experiments of \cite{bolon2013review} and a Random forest classifier is then built on the selected features. For all random forest classifiers, the tree number is set to 500 while the other parameters are set to the default values given by the Scikit-Learn package for Python.

	Similar to our previous work \cite{cao2018improve,cao2018dissimilarity}, a stratified repeated random sampling approach was used to achieve a robust estimate of the performance. The stratified random splitting procedure is repeated 10 times, with 50\% sample rate in each subset. In order to compare the methods, the mean and standard deviations of accuracy are evaluated over 10 runs.

	\section{Results}
	
	The results of mean accuracies, along with the corresponding standard deviation, over the 10 repetitions are shown in Table \ref{tab:51}. GDV and the two static voting methods have almost the same results over the four datasets, but these results are different from the two dynamic weighted voting methods LDV and GLDV. It is not surprising that there is no difference between MVRF and WRF because the datasets we use in this work have only five views, which means that there is no situation like even votes (the worst case would be 3 against 2). Hence as long as there is no extremely big difference among performance of different views, the two static voting methods should have similar results. And the result of GDV confirms our assumption in the previous section that the global weight alone does not contain a lot of personalized information. We can also see that there is a benefit of combining global and local weights as the performance of GLDV is always better than LDV.  From the average ranking value, it can be told that the best method is the proposed GLDV method, followed by GDV. The state of the art solution SVMRFE is ranked at the last place.

	To see more clearly the difference between MCS based methods and SVMRFE, a pairwise analysis based on the Sign test is computed on the number of wins, ties and losses as in the work of \cite{cruz2018dynamic}.  Figure \ref{fig:posthoc3} shows that, when compared to SVMRFE, only the proposed methods LDV and GLDV are significantly better than SVMRFE with  $\alpha$ = 0.10 and 0.05. These results show that the MCS based late integration methods can also be significantly better than the state-of-art Radiomics solutions. 
	
	\begin{minipage}{\textwidth}
		
		\begin{minipage}[b]{0.6\textwidth}
			\centering
			\begin{adjustbox}{max width=1\textwidth}
				\begin{tabular}{ | l  |p{1.5cm}| p{1.5cm}|p{1.5cm}|p{1.5cm}|p{1.5cm}|p{1.5cm}|}
					\hline
					
					Dataset & 
					SVMRFE\newline+RF & MVRF  & WRF&GDV&LDV&GLDV\\ 
					\hline
					nonIDH1&
					
					$76.28\%\newline\pm4.39$
					&
					$82.79\%\newline\pm2.37$
					&
					$82.79\%\newline\pm2.37$&
					$82.79\%\newline\pm2.37$
					&
					$76.98\%\newline\pm1.93$
					&
					$77.44\%\newline\pm2.33$
					\\
					\hline
					IDHcodel&
					
					$73.23\%\newline\pm5.50$
					&
					$76.76\%\newline\pm2.06$
					&
					$76.76\%\newline\pm2.06$
					&
					$76.76\%\newline\pm2.06$
					&
					$74.11\%\newline\pm1.17$
					&
					$74.41\%\newline\pm1.34$
					\\
					\hline
					lowGrade&
					
					$62.55\%\newline\pm3.36$
					&
					$64.41\%\newline\pm3.76$
					&
					$64.41\%\newline\pm3.76$
					&
					$64.41\%\newline\pm3.76$
					&
					$64.41\%\newline\pm3.45$
					&
					$66.05\%\newline\pm3.32$
					\\
					\hline
					progression&
					$62.36\%\newline\pm3.73$
					&
					$61.31\%\newline\pm4.25$
					&
					$61.31\%\newline\pm4.25$
					&
					$61.57\%\newline\pm4.27$
					&
					$62.63\%\newline\pm4.37$
					&
					$62.89\%\newline\pm4.62$
					\\
					\specialrule{.2em}{.1em}{.1em} 
					Average Rank &5.250    &    3.250  &      3.250       & 2.875    &    3.875     &   \textbf{2.500 }
					\\
					\hline
				\end{tabular}
			\end{adjustbox}
			\captionof{table}{Experiment results with 50\% training data 50\% test data for Radiomics data}
			\label{tab:51}    
		\end{minipage}
		\hfill
		\begin{minipage}[b]{0.35\textwidth}
			
			\centerline{\includegraphics[width=1\textwidth]{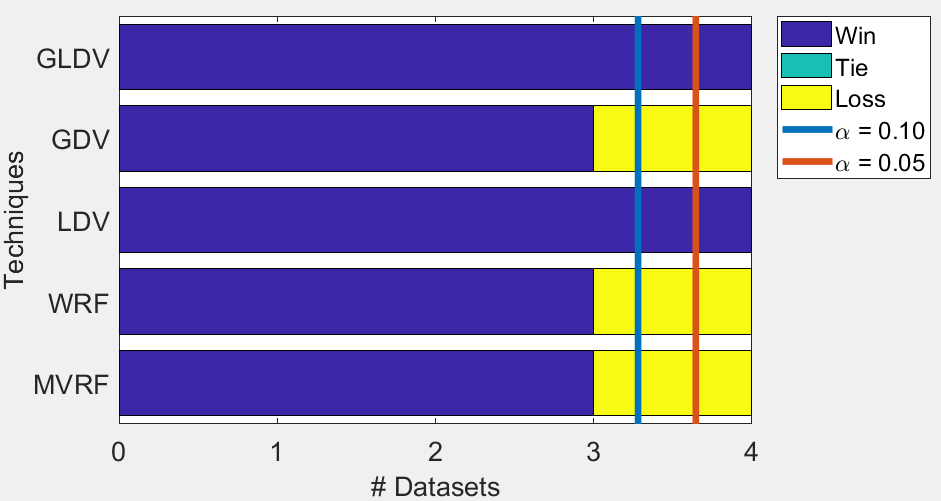}}

			\captionof{figure}{Pairwise comparison between MCS  solutions and SVMRFE. The vertical lines illustrate the critical values considering a confidence level $\alpha$= \{0.10 , 0.05\}.   }
			\label{fig:posthoc3}
		\end{minipage}
		
	\end{minipage}

	When we compare GDV, LDV and GLDV, it can be seen that for nonIDH1 and IDHCodel data, the performance of GLDV is between LDV and GDV (LDV is the worst while GDV is the best). However for the two other datasets, GLDV is always better than both LDV and GDV, which means that for different datasets, the best combination of LDV and GDV should be different. To further study the preference of global weight $ W_{global}$ and local weight $ W_{local}$ for different datasets,  a new combination is formed as:
	
	\begin{equation}\label{gln}
	{W}^{(q)}_{GLnew} = ({W_{global}^{(q)})^{1-a}}\times (W_{local}^{(q)})^{a}
	\end{equation}
	
	From Equation \ref{gln} it can be told that when $a=1$, the combination is only affected by local accuracy while when $a=0$ the combination is only affected by global accuracy. The results of ${W}^{(q)}_{GLnew}$ are shown in Table \ref{extrares}, from which we can confirm our conclusion that for IDHCodel1 and nonIDH data, they get better results when they use more global weight. For lowGrade and progression data, they get better results when they use more local weight.
	
	\begin{table*}[htbp]
		\centering
		\caption{The results of new combinations ${W}^{(q)}_{GLnew}$ with different $a$ value.}
		\label{extrares}
		
		\begin{center}
			\begin{adjustbox}{max width=1\textwidth}
				\begin{tabular}{ | l  | p{1.2cm}| p{1.2cm}| p{1.2cm}| p{1.2cm}|p{1.2cm}| p{1.2cm}|p{1.2cm}| p{1.2cm}| p{1.2cm}|p{1.2cm}| p{1.2cm}|}
					\hline
					Dataset     & a=0 (GDV)          & a=0.1            & a=0.2            & a=0.3            & a=0.4            & a=0.5            & a=0.6            & a=0.7            & a=0.8            & a=0.9            & a=1 (LDV)          \\ \hline
					nonIDH      & $\textbf{82.79}\%\newline\pm2.37$ & $\textbf{82.79}\%\newline\pm2.37$ & $\textbf{82.79}\%\newline\pm2.37$ & $82.32\%\newline\pm2.13$ & $81.16\%\newline\pm3.02$ & $80.23\%\newline\pm2.80$ & $79.99\%\newline\pm3.15$ & $79.30\%\newline\pm2.42$ & $77.90\%\newline\pm2.38$ & $77.44\%\newline\pm2.33$ & $76.97\%\newline\pm1.93$ \\ \hline
					
					IDHCodel1   & $\textbf{76.76}\%\newline\pm2.06$ & $\textbf{76.76}\%\newline\pm2.06$ & $\textbf{76.76}\%\newline\pm2.06$ & $75.88\%\newline\pm1.76$ & $75.58\%\newline\pm1.34$ & $75.29\%\newline\pm1.44$ & $75.29\%\newline\pm1.44$ & $75.29\%\newline\pm1.95$ & $75.00\%\newline\pm1.97$ & $75.00\%\newline\pm1.97$ & $74.41\%\newline\pm1.34$ \\ \hline
					
					lowGrade    & $64.41\%\newline\pm3.75$ & $64.41\%\newline\pm3.75$ & $64.41\%\newline\pm3.75$ & $\textbf{64.65}\%\newline\pm3.57$ & $64.41\%\newline\pm3.45$ & $64.41\%\newline\pm3.45$ & $\textbf{64.65}\%\newline\pm3.72$ & $64.18\%\newline\pm4.18$ & $63.48\%\newline\pm3.75$ & $63.48\%\newline\pm3.45$ & $63.95\%\newline\pm3.64$ \\ \hline
					
					progression & $61.57\%\newline\pm4.27$ & $61.57\%\newline\pm4.27$ & $61.84\%\newline\pm3.57$ & $62.10\%\newline\pm3.56$ & $62.36\%\newline\pm3.91$ & $62.10\%\newline\pm4.43$ & $62.36\%\newline\pm4.41$ & $\textbf{63.42}\%\newline\pm4.62$ & $62.89\%\newline\pm4.77$ & $62.89\%\newline\pm4.77$ & $62.36\%\newline\pm4.56$ \\ \hline
				\end{tabular}    
			\end{adjustbox}
		\end{center}
		
	\end{table*}
	In general, all MCS based late integration methods are better than feature selection methods. Majority voting is simple and efficient. GLDV is only better than majority voting on two datasets. But LDV and GLDV are preferable for Radiomics applications in the following three ways: (i) they give different weights of each view to each test sample, so that each test sample uses a different combination of classifiers to give a personalized decision; (ii) they are significantly better than the state of art work in Radiomics; (iii) the performance of GLDV can be further improved by adjusting the proportion of local weight and global weight. Note that other parameters like the neighborhood size can also be adjusted to optimize the performance. Compared to static voting, the disadvantage of dynamic voting is that it is more complex and less efficient.

	\section{Conclusions}
	In the state of art works of Radiomics, most studies used feature selection methods as a solution for the HDLSS problem. In this work, we have treated Radiomics as a multi-view learning problem and investigated the potential of MCS based late integration methods, proposed earlier in \cite{cao2018dissimilarity}. In particuler, we have investigated some dynamic voting based MCS methods, that can give each patient a personalized prediction by dynamically integrating the classification result from each view. We believe these methods have a great potential and can significantly outperform early integration methods that make use of feature selection in the concatenated feature space. 
	
	To confirm our hypothesis, a representative early integration method, five MCS methods including three dynamic voting methods and two static voting methods, have been compared on four Radiomics datasets.  We conclude from our experiments that all MCS based late integration methods are generally better than the state of art Radiomics solution, but only LDV and GLDV are significantly better, which shows the potential of MCS based late integration methods of being a better solution than the state-of-art Radiomics solutions. 

	\section*{Acknowledgment}
	This work is part of the DAISI project, co-financed by the European Union with the European Regional Development Fund (ERDF) and by the Normandy Region.

	\bibliographystyle{ieeetr}
	\bibliography{sample}
	
\end{document}